\documentclass[12pt]{article}

\pdfoutput=1

\usepackage[english]{babel}
\usepackage[utf8]{inputenc} 
\usepackage[T1]{fontenc}

\usepackage[letterpaper,top=2.54cm,bottom=2.54cm,left=2.5cm,right=3cm,marginparwidth=1.75cm]{geometry}

\usepackage[english]{babel}
\usepackage[style=ieee,backend=bibtex]{biblatex}
\usepackage{subcaption}
\usepackage{amsmath}
\usepackage{csquotes}
\usepackage{float}
\usepackage{hyperref}

\hypersetup{
    colorlinks=true,
    linkcolor=blue,
    filecolor=magenta,      
    urlcolor=cyan,
    citecolor={blue},
}

\usepackage{bm}
\usepackage{xargs}
\usepackage{titling}
\usepackage{placeins}
\usepackage{setspace}
\usepackage{booktabs}
\usepackage{siunitx}
\usepackage{physics}
\usepackage{braket}
\usepackage{tikz}
\usetikzlibrary{quantikz}

\usepackage{graphicx}
\graphicspath{{thesis-figs/}}

\usepackage{etoolbox}
\patchcmd{\thebibliography}{\section*{\refname}}{}{}{}

\usepackage{xcolor}

\newcommand{\qcsim}{{$|\braket{\phi (\vec x_j) \vert \phi (\vec x_i)}|^2$ }}

\newcommand{\code}{\texttt}

\usepackage[markup=underlined]{changes}
\makeatletter
\@namedef{Changes@AuthorColor}{magenta}
\colorlet{Changes@Color}{magenta}
\makeatother
\usepackage{authblk}

\begin{document}

\begin{titlepage}
   \begin{center}

        \vspace*{0.5cm}
        
        \Large ENPH 455 Final Report

        \vspace{0.5cm}
        
        \Huge Design and Implementation of a Quantum Kernel for Natural Language Processing
            
        \vspace{1.5cm}
        \normalsize by
        \vspace{1.25cm}
        
        \Large  Matt Wright
        
        \vspace{0.5cm}
        
        \normalsize Supervisor: Dr. Stephen Hughes
        
        \vfill \normalsize
        
        A thesis submitted to the \\
        Department of Physics, Engineering Physics, and Astronomy \\
        in conformity with the requirements for \\
        the degree of Bachelor of Applied Science \\
            
        \vspace{0.8cm}
        
        Queen’s University \\
        Kingston, Ontario, Canada \\
        April 2022
        
        \vspace{1cm}
        
        Copyright \copyright \ Matt Wright, 2022
            
   \end{center}
\end{titlepage}

% \maketitle
% \thispagestyle{empty}

\pagebreak
\pagenumbering{roman}
\setcounter{page}{1}

\section*{Abstract}
\begin{spacing}{1}
%No references, define all acronyms
%Goals, constraints, results

% 1. context or background information
% 2.  central questions or statement of the problem your research addresses
% 3. previous work
% 4. goal
% 5. methods
% 6. findings, results
% 7. the significance or implications of your findings

% Your thesis should include either an Abstract or an Executive Summary.  The abstract is often single-spaced, but it may be double-spaced if you prefer.  It should not be more than 400 words.  The abstract does not count towards the word count of the main body of your report.  Abstracts should stand alone – that is, should not include references, should define all acronyms, and should describe the main goals, constraints and results of the thesis.  

% Background & motivation
Natural language processing (NLP) is the field of computer science that attempts to make human language accessible to computers, and it relies on developing a mathematical model to express the meaning of symbolic language. One such model, referred to as DisCoCat, defines how to express both the meaning of individual words as well as the compositional nature of the sentence specified by grammar. This category-theoretic model is costly to implement on a classical computer but has an equivalent quantum mechanical formulation, which has lead to its implementation on quantum computers and the creation of the field quantum NLP (QNLP). This recent experimental work used quantum machine learning techniques to learn a mapping from text to class label using the expectation value of the quantum encoded sentence. Some theoretical work has been done on computing the similarity of sentences using this quantum model but they rely on an unrealized quantum memory store.
% Goals and constrain
The main goal of this thesis is to leverage the DisCoCat model to design a quantum-based kernel function that can be used by a support vector machine (SVM) for NLP tasks. The kernel function must compute the similarity of two given sentences with high accuracy, efficiency, and resilience to noise.
% Results and recommendations
Two similarity measures were studied: (i) the transition amplitude approach and (ii) the SWAP test. A simple NLP meaning classification task from previous work was used to train the word embeddings and evaluate the performance of both models. The Python module \code{lambeq} and its related software stack was used for implementation. The explicit model from previous work was used to train word embeddings and achieved a testing accuracy of $93.09 \pm 0.01$\%. It was shown that both the SVM variants achieved a higher testing accuracy of $95.72 \pm 0.01$\% for approach (i) and $97.14 \pm 0.01$\% for (ii). The SWAP test achieved a slightly higher accuracy and yielded a more appropriate similarity matrix. The SWAP test was then simulated under a noise model defined by the real quantum device, \code{ibmq\_guadalupe}. The explicit model achieved an accuracy of $91.94 \pm 0.01$\% while the SWAP test SVM achieved 96.7\% on the testing dataset, suggesting that the kernelized classifiers are resilient to noise. These are encouraging results and motivate further investigations of our proposed kernelized QNLP paradigm.

\vspace{1cm}
\section*{Acknowledgements}
I would like to thank my supervisor, Dr. Stephen Hughes, for his insight and guidance throughout this project. This field was new to both of us so I appreciate him for wondering into the unknown with me. 

As this is the culmination of my undergraduate degree, I would also like to thank my family, friends, peers, and the Queen's Faculty that have all supported me and played a part in the development of my personal and professional self.

\end{spacing}
\pagebreak

%Main Body less than 5000 words

\tableofcontents

\pagebreak

\listoffigures

% comment out for now
\pagebreak
\listoftables

\pagebreak
\cleardoublepage\pagenumbering{arabic}

\section{Introduction and Motivation}

% look at problem def document

% Talk about NLP
% Classical strategies for NLP
% Shortcomings
% DISCOCAT
% Quantum implementatoin of discocat
% shortcomings
% "The goal of this thesis ..."

Machine learning (ML) has applications in almost all computational tasks and one field under active research is natural language processing (NLP), which attempts to make human language accessible to computers \cite{eisensteinIntroductionNaturalLanguage2019}. NLP research is motivated in part by the prevalence of natural language data in the form of emails, tweets, books, and more -- effectively leveraging this data would be incredibly valuable. Applications of NLP include behavioural analysis, spam filtering, virtual assistants, and more. To understand natural language, one must understand the ambiguities, context, and the large set of complex linguistic rules that define the meaning of language. Consequently, developing a mathematical model for language that computers can efficiently implement is a difficult problem and is at the core of NLP research.
Most common approaches to NLP are limited to analyzing either the meaning of the individual words and phrases  based on context or feeding the text into large black-box neural networks. Alternatively, some approaches leverage the grammatical structure of the text. Unifying these two approaches is an active area of research \cite{lenciDistributionalModelsWord2018} and is the focus of this project as it often leads to an exponential increase in the size of the vector space required to describe a given sentence \cite{coeckeFoundationsNearTermQuantum2020}. The tensor network structure, (more precisely, the category-theoretic nature \cite{coeckeMathematicalFoundationsCompositional2010}) of some of these grammar models, namely \emph{DisCoCat}, has motivated the recent application of quantum computing to NLP \cite{coeckeFoundationsNearTermQuantum2020, zengQuantumAlgorithmsCompositional2016, meichanetzidisGrammarAwareQuestionAnsweringQuantum2020, lorenzQNLPPracticeRunning2021}, creating a field dubbed \emph{quantum natural language processing} (QNPL).

Quantum information science (QIS) has become a rich and vibrant area of research because of the fundamental challenges quantum information theory poses to the classical computational paradigm, together with experimentally realizable systems \cite{nielsenChuang}. The interest in quantum computing is motivated by the proven exponential algorithmic complexity speed-ups in addition to the high-dimensional computational vector spaces that these computers can access. This has lead to the application of quantum computing to areas like cryptography, chemistry, machine learning, and  NLP.

Quantum hardware today is in the so-called noisy intermediate-scale quantum (NISQ) device era. As a result, many of the famed quantum algorithms that promise exponential speed-ups are not attainable at large scales yet, which has lead to creation of a class of hybrid classical-quantum algorithms that reduce the size of the quantum circuits. This is the field of \textit{quantum machine learning}. These algorithms have been shown to be equivalent to kernel methods \cite{havlicekSupervisedLearningQuantumenhanced2019, schuldSupervisedQuantumMachine2021}, which have two distinct manifestations: \textit{explicit} and \textit{implicit} models \cite{schuldQMLinFeatureHilbertSpaces}. Explicit models use a variational circuit to learn how to classify the data while implicit models compute the similarity of two data points for use by a kernelized classifier such as a support vector machine (SVM). Thus far, QNLP work has been restricted to building explicit models with the exception of some theoretical work using nearest neighbour classifiers but this is outdated and can be inefficient.

The goal of this thesis is to design a high accuracy, robust implicit quantum machine learning model that leverages the DisCoCat grammar model to encode natural language data. This will require the design and analysis of different algorithms to measure the fidelity of these quantum states to determine the similarity of two sentences.

\section{Background Theory}

\subsection{Grammatical Model of Meaning: DisCoCat}
\label{sec:discocat}

\begin{figure}[h]
    \centering
    \subfloat[Simple noun, transitive verb, noun sentence,  from~\cite{coeckeFoundationsNearTermQuantum2020}.]{\includegraphics[width=.4\linewidth]{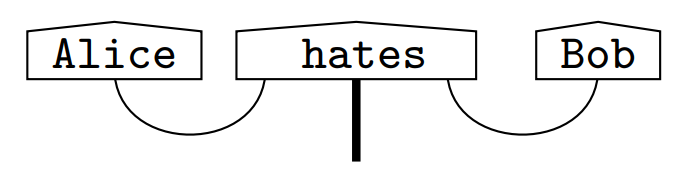}} \\
    % \vspace{-2mm}
    \subfloat[More complex sentence, from \cite{coeckeFoundationsNearTermQuantum2020}.]{\includegraphics[width=0.9\linewidth]{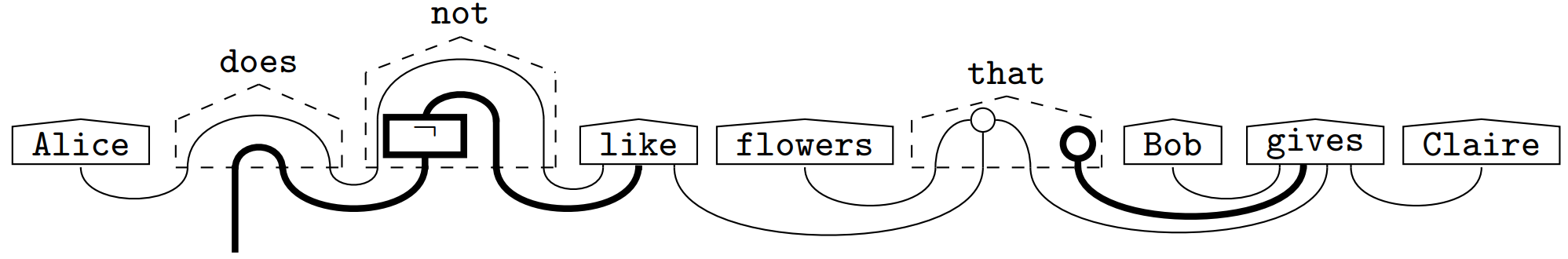}}    
    \vspace{-2mm}
    \caption[Diagrammatic representation of two DisCoCat encoded example sentences.]{The two subfigures above show the diagrammatic representation of two sentences using DisCoCat. Mathematically, the boxes (like \texttt{Alice}) are tensors and the strings connecting the boxes are reduction operations that are derived from the grammatical structure of the sentence. We will see shortly that in QNLP, the boxes are quantum states and the wires are quantum operations that typically involve entangling.}
    \label{fig:discocat-ex}
\end{figure}

One of the main complexities that the field of NLP continues to deal with is building a mathematical representation of natural language. Historically, the approach to this representation has been based on the paradigm famously summarized by Firth that, ``you shall know a word by the company it keeps'' \cite{firth}. This states that the meaning of words can be learned by examining their context in a corpus, which is a \textit{distributional} approach to NLP most commonly employed by bag-of-words models \cite{harrisDistributionalStructure, lenciDistributionalModelsWord2018}. Modern ML approaches have moved away from this paradigm but still are limited in their ability to incorporate the grammatical and linguistic structure of sentences \cite{lenciDistributionalModelsWord2018} -- this is what \textit{DisCoCat} aims to solve. DisCoCat is a grammar model that expresses both the meaning of words (\textit{\underline{dis}tributional}) and how words interact (\textit{\underline{co}mpositional}) to form a sentence, and is based on a \textit{\underline{cat}egory-theoretic} model. For a deeper examination of the sophisticated mathematics underpinning this model, we refer the reader to the topics of applied category theory, categorical quantum mechanics, and ZX-calculus \cite{csc, coeckeFoundationsNearTermQuantum2020, coecke_kissinger_2017}.

Words in DisCoCat are categorized as types specified by a pregroup grammar \cite{lambek2008}, which defines the order of tensor used to represent them and the reduction rules for combining them \cite{lorenzQNLPPracticeRunning2021}. The pregroups used here are nouns, $n$, and sentences, $s$. Together, the tensors and reduction rules have the effect of modelling the flow of meaning through the sentence \cite{coeckeFoundationsNearTermQuantum2020}. This model has a natural diagrammatic representation as shown in Fig.~\ref{fig:discocat-ex}. To combine multiple words, we must combine the individual vector spaces of each word using tensor products.

\subsection{Quantum Computing}
% fundamentals
% TODO: post-selection?

Quantum computing is a branch of QIS that leverages the phenomenons of superposition and entanglement to perform computation. The basic unit of information is a quantum bit or \emph{qubit}, which can be represented as an abstract mathematical object that lives in a two-dimensional complex Hilbert space \cite{nielsenChuang}. We assign this two-level quantum system the orthonormal basis vectors $\ket{0}$ and $\ket{1}$, which correspond to the measurement outcomes ``0'' and ``1'', respectively. One can represent the \textit{state} of the system as a linear combination of these basis vectors as,
\begin{equation}
    \ket{\psi} = \alpha\ket{0} + \beta\ket{1}
    \label{eq:general_state},
\end{equation}
where $\alpha$ and $\beta$ are complex coefficients that satisfy $|\alpha|^2 + |\beta|^2=1$. The linear combination of states is referred to as \textit{superposition} and is one of the oddities of quantum theory because the system (the qubit in our case) can exist in all possible states at once. This leads to the  probabilistic nature of quantum mechanics that arises from measuring states in superposition but only observing one of the outcomes 0 or 1. The Born rule is the postulate of quantum mechanics that specifies that the probability of measuring some outcome, $i$ (where $i=0,1$ for a qubit), is given by the projection of the system's state onto the subspace spanned by the eigenvectors with eigenvalues $i$:  $P[i] = |\braket{i | \psi}|^2$. It can then be seen that the probabilities associated with the outcomes specified by the general qubit state in Eq.~\eqref{eq:general_state} are $P[0]=|\alpha|^2$ and $P[1]=|\beta|^2$. The state of a system is evolved by the operation of a \textit{unitary transformation}, $\widehat U$, which is represented as $\ket{\psi_1} = \widehat U \ket{\psi_0}$.

To represent the composite space of multiple qubits, one must use \emph{tensor products} to combine the individual Hilbert spaces. That is, the joint state of two uncorrelated qubits is represented as $\ket{\psi} = \ket{\psi_1} \otimes \ket{\psi_2}$. Thus, an $n$-qubit system has a dimension of $2^n$. Representing a general state vector in this system on a classical computer requires an exponential amount of memory, which quickly becomes intractable. One represents a \textit{quantum circuit} by combining $n$ qubits and applying single- or multi-qubit operations (\textit{gates}) to the system.  

For a more exhaustive and in-depth introduction to the field of QIS, we refer the reader to Ref.~\cite{nielsenChuang}.

\subsection{Quantum Machine Learning}
% variational circuits
% implicit and explicit
% Kernels and SVM
% similarity measures

\begin{figure}[h]
  \centering
  \includegraphics[width=.5\textwidth]{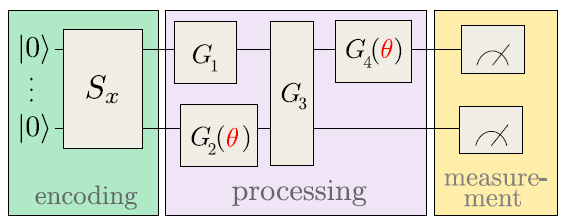}  
  \caption[General parameterized quantum circuit architecture.]{Parameterized quantum circuit architecture for a data point, $x$, and set of parameters, $\theta$. Encoding uses a feature map to transform classical data into a quantum state and the parameterized ``processing'' operations can be optimized to learn a desired mapping -- this is an explicit model. One can see the similarity between such a circuit and a classical machine learning model. Taken from \cite{schuldSupervisedQuantumMachine2021}.}
  \label{fig:param-circuit-components}
\end{figure}

While some near-term quantum devices have shown increasingly promising results, they are not fault-tolerant and suffer from decoherence due to noise. Therefore, so-called \emph{variational quantum algorithms} have become a vibrant area of research because they use a hybrid quantum-classical approach wherein parameterized quantum circuits are executed on quantum computers and the parameters are optimized using classical optimization techniques. This keeps the classically intractable portion of the computation task on the quantum computer while reducing the depth of the circuit. This class of hybrid quantum-classical algorithms is closely related to the field of \emph{quantum machine learning} (QML) since many techniques of classical machine learning are applied here and the quantum circuit is, in essence, learning a function. QML algorithms can have three main components shown in Fig.~\ref{fig:param-circuit-components}: (i) the data encoding/embedding layer, (ii) the processing layer, and (iii) the measurement. The encoding layer ingests the data and converts it to a quantum state representation through a \emph{feature map}. The processing layer is a parameterized operation typically composed of rotating and entangling gates and is subject to training. The measurement specifies which qubits to measure and in what basis, and the outcome of which is interpreted as the prediction. This type of architecture will be referred to as an \emph{explicit} QML model \cite{schuldQMLinFeatureHilbertSpaces} as the circuit directly learns how to transform and separate data.

A recent argument showed that QML algorithms can be equivalently described as \emph{kernel methods} \cite{schuldSupervisedQuantumMachine2021, schuldQMLinFeatureHilbertSpaces}, which are pervasive in ML. Many classical ML algorithms use a (classical) \textit{feature map} to transform data to a more rich \emph{feature space} in which the solution is easier to learn. %A simple but illustrative example of the benefits of a feature map can be seen in Fig.~\ref{fig:feature-map}. 
Kernel methods are a different class of algorithms that use a specified \textit{kernel function} to compute the similarity of a pair of data points. Kernel functions are desirable as they can implicitly map data to the rich feature space without explicitly representing the vectors in this space,  allowing for the algorithms to compute similarity in, say, infinite-dimensional feature spaces \cite{bishopMlBook}.
Similarly, \textit{quantum kernels} use quantum feature maps to transform data to quantum states and compute similarity by estimating the inner-product, $|\braket{\psi | \phi}|^2$. The use of such kernels is motivated by their ability to efficiently compute inner-products in exponentially high dimensional Hilbert spaces. For example, consider a pair of data points $\vec x_1$ and $\vec x_2$, and a classical feature map $\vec \phi (\vec x)$. Classically, the kernel could be given by,
\begin{equation} \label{eq: classical-kernel}
    k(\vec x_1, \vec x_2) = \vec \phi (\vec x_2)^T \vec \phi (\vec x_1).
\end{equation}

In the quantum realm, the feature map uses an operator parameterized by a given data point: $\widehat U_{\phi (\vec x)} \vert 0^{\otimes n} \rangle = \vert \phi (\vec x) \rangle$, and the kernel computes the inner-product of the resulting states,
\begin{align} \label{eq:quantum-kernel}
    k(\vec x_1, \vec x_2) 
        &= |\braket{\phi (\vec x_2) \vert \phi (\vec x_1)}|^2. %\nonumber \\
        %&= \braket{0^{\otimes n} \vert \hat U_{\phi (\vec x_1)} ^ \dagger \hat U_{\phi (\vec x_2)}  \vert 0^{\otimes n}},
\end{align}

With this kernel function one can the build a classification model using, such as a \emph{support vector machine} (SVM), which is trained on the Gram matrix of the dataset, $K_{i, j} = k(\vec x_i, \vec x_j)$. This type of QML model is an \emph{implicit} model since the quantum computer does not map directly to the model output \cite{schuldQMLinFeatureHilbertSpaces}. An SVM is a \emph{sparse maximum-margin} machine learning model that finds a hyper-plane defined by a subset of data points, called \emph{support vectors}, to separate training data \cite{bishopMlBook}. An SVM is \textit{sparse} because it only requires the support vectors for prediction as they alone define the decision boundary.  The model is referred to as \textit{maximum-margin} because the objective of training is to find the hyper-plane that maximizes the distance between the decision boundary and the training data points that are closest to it. Since dot products (more generally, vector similarities) are fundamental to SVMs, kernels are often used to implicitly enhance data and have become a promising framework for QML \cite{rebentrostQuantumSupportVector2014, havlicekSupervisedLearningQuantumenhanced2019}. In quantum SVMs, the quantum computer is responsible for only the similarity computation which is used by the classical computer to determine the model output.

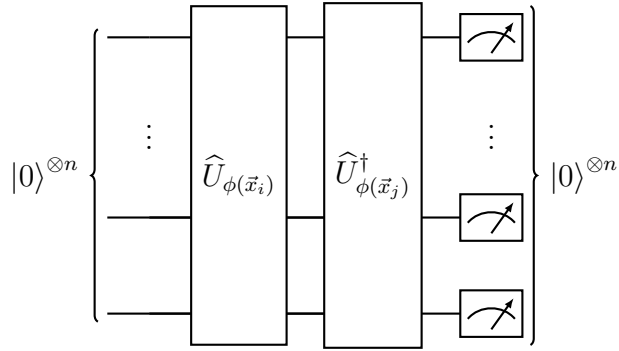
\begin{figure} [h]
    \centering
    \begin{quantikz}
        \lstick[wires=4]{$\ket{0}^{\otimes n}$} & \qw & \gate[wires=4, nwires=2]{\widehat U_{\phi (\vec x_i)}} & \gate[wires=4,nwires=2]{\widehat U_{\phi (\vec x_j)}^\dagger}  &   \meter{}  \rstick[wires=4]{$\ket{0}^{\otimes n}$} \\
        & \vdots &&& \vdots \\
        & \qw & \qw & \qw  & \meter{} \\
        & \qw & \qw & \qw  & \meter{}
    \end{quantikz}
    \caption[Transition amplitude similarity measure in quantum circuit form.]{Circuit diagram for a transition amplitude similarity measure. The circuit is generalized for $n$ qubits that are encoded with the quantum feature map $\widehat U_{\phi (\vec x)}$. The probability of measuring all zeros is used to determine the fidelity of the two quantum states corresponding to $\vec x_i$ and $\vec x_j$ via Eq.~\eqref{eq:transition-amp-kernel}.}
    \label{fig:trans-amp-circ}
\end{figure}

To implement a quantum kernel, there are two main approaches we discuss for estimating the fidelity of two quantum states. One could use the \textit{transition amplitude} approach wherein the quantum feature map, $\widehat U_{\phi (\vec x_i)}$, encoding $\vec x_i$ is applied to the $\ket{0^{\otimes n}}$ state, followed by the adjoint of the operator, $\widehat U_{\phi (\vec x_j)}^\dagger$, for data point $x_j$. This circuit is then executed for many ``shots''  to estimate the probability of measuring $\ket{0^{\otimes n}}$, which estimates,
\begin{equation} \label{eq:transition-amp-kernel}
    |\braket{\phi (\vec x_j) \vert \phi (\vec x_i)}|^2 = |\braket{0^{\otimes n} |\widehat U_{\phi (\vec x_j)}^\dagger  \widehat U_{\phi (\vec x_i)} | 0^{\otimes n}}|^2,
\end{equation}
which resembles the familiar quantum mechanical transition amplitude of the operator. The quantum circuit diagram for such a circuit is given in Fig.~\ref{fig:trans-amp-circ}.

For a geometric argument of this approach, one must recall that quantum gates can be represented as rotations of the system's state along the surface of the Bloch sphere \cite{nielsenChuang}. So one can think of $\widehat U_{\phi (\vec x_i)}$ as a rotation by an amount specified by $\vec x_i$ and, conversely, $\widehat U_{\phi (\vec x_j)}^\dagger$ is a rotation in the opposite direction by $\vec x_j$. Thus, the similarity function defined in Eq.~\eqref{eq:transition-amp-kernel} is measuring how close the system is to its initial state after being rotated in the forward direction by $x_i$ and then rotated in the reverse direction by $x_j$.

\begin{figure}[h]
    \centering
    \begin{quantikz}
        \lstick{$\ket{0}$} & \gate{H} & \ctrl{1} & \gate{H}\slice{$\ket{\Psi}$} & \meter{} \\
        \lstick{$\ket{\phi (\vec x_i)}$} & \qw & \gate[wires=2]{{\rm SWAP}} & \qw & \qw \\
        \lstick{$\ket{\phi (\vec x_j)}$} & \qw & & \qw & \qw
    \end{quantikz}
    \caption[SWAP test circuit diagram.]{Circuit diagram for a SWAP test similarity measure. The $H$ refers to the Hadamard gate and the multi-qubit gate is the controlled-SWAP gate. The top qubit is the ``ancillary'' qubit and the other qubits are prepared in the states corresponding to two data points. The similarity of the two arbitrary states is estimated with this circuit using Eq.~\eqref{eq:swap-test-sim}. The state of $\Psi$ is given in Eq.~\eqref{eq:swap-test-state}.}
    \label{fig:swap-circ}
\end{figure}
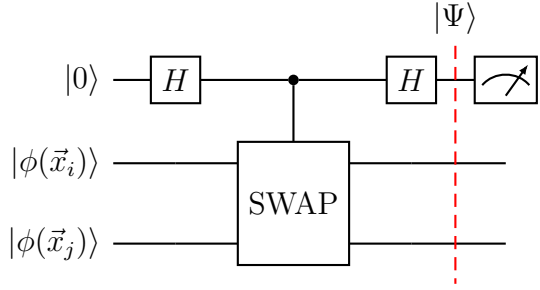

Alternatively, one could use the \textit{SWAP test} as shown in Fig.~\ref{fig:swap-circ} to estimate the fidelity of the quantum states using an extra ``ancillary'' qubit \cite{havlicekSupervisedLearningQuantumenhanced2019, buhrmanQuantumFingerprinting2001}. The SWAP test uses a controlled-SWAP gate exchange two states (i.e., $\ket{\phi}\ket{\psi} \rightarrow \ket{\psi}\ket{\phi}$) controlled on the ancillary qubit, which is put in and then taken out of superposition using the  Hadamard gate. Here, we use the quantum encoded data points as the two states.

Before measurement, the final state of the circuit shown in Fig.~\ref{fig:swap-circ} is given by \cite{buhrmanQuantumFingerprinting2001},
\begin{equation} \label{eq:swap-test-state}
    \ket{\Psi} = \frac{1}{2} \ket{0} (\ket{\phi (\vec x_i)}\ket{\phi (\vec x_j)} + \ket{\phi (\vec x_j)}\ket{\phi (\vec x_i)}) + \frac{1}{2} \ket{1} (\ket{\phi (\vec x_i)}\ket{\phi (\vec x_j)} - \ket{\phi (\vec x_j)}\ket{\phi (\vec x_i)}).
\end{equation}

It can then be seen (using the Born rule) that measuring 0 for the ancillary qubit has the probability  $Pr[0] = \frac{1}{2} + \frac{1}{2} |\braket{\phi (\vec x_j) \vert \phi (\vec x_i)}|^2$. Thus, repeating this experiment many times allows us to obtain an estimate for $Pr[0]$, which can then be used to compute the similarity of the data points via,
\begin{equation} \label{eq:swap-test-sim}
    |\braket{\phi (\vec x_j) \vert \phi (\vec x_i)}|^2 = 2Pr[0] - 1.
\end{equation}

Since the Hadamard  gate, $H$, is its own inverse (i.e., $\widehat H \widehat H = \widehat I$),  if there was no controlled-SWAP gate then the ancillary qubit would always be 0. Intuitively, one can imagine that the similarity of the two states in the circuit above is measured based on the amount of disturbance that swapping the two states has on the ancillary qubit's superposition. This can easily be generalized to multi-qubit states using additional ancillary qubits or just additional controlled-SWAP gates targeted on different qubit pairs.

\subsection{Quantum Natural Language Processing}
% Quantum implementation of discocat
% related works

% \begin{figure}[h]
%     \centering
%     \subfloat[DisCoCat representation of a sentence with pregroup type labelled.]{\includegraphics[width=.4\linewidth]{discocat-sentence.png}} \hspace{5mm}
%     \subfloat[Quantum circuit representation of the sentence in (a). Boxes (iii), (i), (ii), (v) correspond to ``person'', ``prepares'', ``tasty'', ``dinner'', respectively. (iv) is the quantum operation (Bell measurement) that corresponds to the ``cup'' (reduction operation, $\cup$) joining ``prepares'' and ``tasty''.]{\includegraphics[width=0.45\linewidth]{discocat-qc.png}} 
%     \caption[QNLP example using DisCoCat.]{Both figures taken from \cite{lorenzQNLPPracticeRunning2021} to exemplify how a sentence is converted from DisCoCat string diagram (a) to a quantum circuit (b).}
%     \label{fig:discocat-qc}
% \end{figure}

\begin{figure}[h]
    \centering
    \begin{subfigure}[b]{\linewidth}
    \centering
        \includegraphics[width=.4\linewidth]{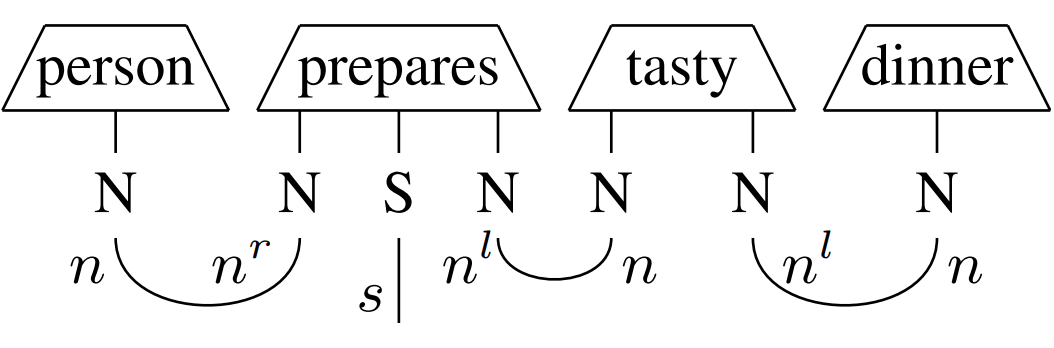}
        \caption{DisCoCat string diagram representation of a sentence with pregroup type labelled.}
    \end{subfigure}
    
    \begin{subfigure}[b]{\linewidth}
    \centering
        \includegraphics[width=0.45\linewidth]{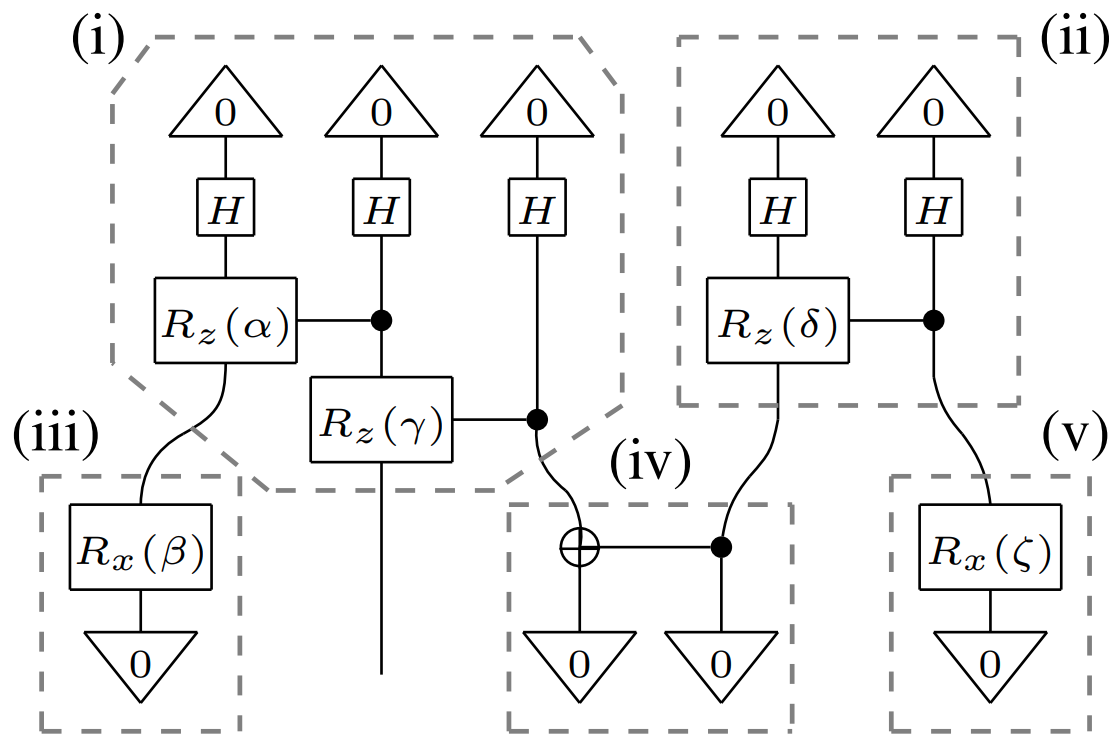}
        \caption{Quantum circuit representation of the sentence in (a). Boxes (iii), (i), (ii), (v) correspond to ``person'', ``prepares'', ``tasty'', ``dinner'', respectively. (iv) is the quantum operation (Bell measurement) that corresponds to the ``cup'' (reduction operation, $\cup$) joining ``prepares'' and ``tasty''. The triangles with 0 at the bottom of the circuit indicate post-selecting for 0 measurements.}
    \end{subfigure}
    
    \caption[QNLP example using DisCoCat.]{Both figures taken from \cite{lorenzQNLPPracticeRunning2021}, exemplify how a sentence is converted from DisCoCat string diagram (a) to a quantum circuit (b). Note that $q_n=1=q_s$.}
    \label{fig:discocat-qc}
\end{figure}

The tensor network nature of DisCoCat (as described in Section~\ref{sec:discocat}) means that it requires an intractable amount of resources to model modest sized NLP problems on a classical computer; however, it is naturally and efficiently realizable on a quantum computer. In fact, the model's origin has roots in both linguistics and categorical quantum mechanics \cite{csc, coeckeMathematicalFoundationsCompositional2010}. In this formulation, the ``boxes'' that represent word tensors are quantum states of qubits and the ``strings'' connecting boxes are implemented as quantum entangling operations as seen in Fig.~\ref{fig:discocat-qc}. This has lead to the creation of a new field named quantum natural language processing (QNLP).

\begin{figure}[h]
    \centering
    \includegraphics[width=0.5\linewidth]{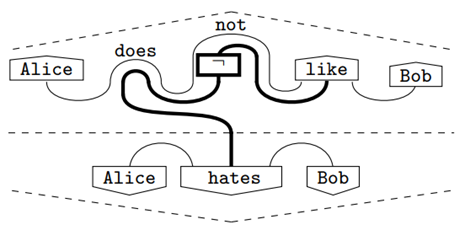}
    \caption[Diagrammatic bra-ket inner-product of two sentences using DisCoCat.]{Diagrammatic bra-ket inner-product of two sentences using DisCoCat  string diagrams. The noun pregroup type corresponds to thin strings and the bold strings correspond to the sentence group. Also note that some simplifications were made by removing the words ``does'' and ``not'' for efficiency.}
    \label{fig:inner-p}
\end{figure}

Existing experimental QNLP work \cite{lorenzQNLPPracticeRunning2021, meichanetzidisGrammarAwareQuestionAnsweringQuantum2020} use DisCoCat to represent sentences as parameterized quantum circuits. This allows the practitioner to define how words are represented by specifying the number of qubits per $n$ and $s$ pregroups ($q_n$ and $q_s$, respectively) and the circuit structure used to encode meaning (i.e., the \textit{ansatz}). These circuits are trained on a dataset to learn the quantum state representation of each word (i.e., the \textit{word embeddings}) using a hybrid quantum-classical optimization approach. Model prediction is done using the explicit approach of measuring the output qubit(s) of a given circuit. These qubits correspond to the quantum representation of the sentences (see the free thick black wires at the bottom of Fig.~\ref{fig:discocat-ex} or the $s$ string in Fig.~\ref{fig:discocat-qc}). This variational quantum encoding requires \textit{post-selection} (shown in Fig.~\ref{fig:discocat-qc}(b)) wherein only the circuit execution results that yield 0 for certain qubits are valid representations of the state. Other theoretical work \cite{zengQuantumAlgorithmsCompositional2016, oriordanHybridClassicalquantumWorkflow2020} propose computing the similarity of sentences by evaluating the inner-product of two quantum-encoded sentences. Diagrammatically, this can be seen in Fig.~\ref{fig:inner-p}. This work relies on the use of a quantum random access memory (qRAM) \cite{Giovannetti_2008_QRAM} that is not physically realizable yet and is applied to a nearest-neighbour classifier,  which is inefficient for prediction.

% TODO: better picture?

\section{Problem Definition}
% GOAL: Design a kernelized classification model that extends DisCoCAt
% CONSTRAINTS: Width and depth
% EVALUATION: Accuracy

Linguistic structure is a critical aspect of language comprehension but is often overlooked in classical NLP techniques. Recent experimental work in QNLP is limited to explicit quantum algorithms that compute single-sentence expectation values to obtain a binary label \cite{meichanetzidisGrammarAwareQuestionAnsweringQuantum2020, lorenzQNLPPracticeRunning2021}. Other theoretical models propose leveraging sentence similarity in a nearest neighbour classifier, but this can be inefficient because prediction requires that the similarity between a given data point and all training data samples be computed \cite{zengQuantumAlgorithmsCompositional2016, oriordanHybridClassicalquantumWorkflow2020}. Further, the design of an executable algorithm and experimental analysis of such a model has not been done. 

This thesis project aims to design an algorithm that extends the predominant QNLP grammar model, DisCoCat, by designing a kernel function that leverages the model to allow for more robust and efficient classification algorithms, e.g., SVMs. The main measure that will be used to validate the results of this project is accuracy on a given NLP classification task compared to existing algorithms. The design is constrained primarily by the level of noise in current quantum devices as well as the number of qubits such devices have (say $<$50) so the number of gates (depth) and qubits (width) are the main technical constraints for real world application.

\section{Methodology}  \label{sec:meth}
% RUBRIC: The student compares multiple ways of solving the problem and selects some using evidence based reasons.
% software tools
% simulators?
% similarity function

% other similarity-based classifiers

When considering the software tools available today for designing quantum algorithms, there are many different options but the two classes of tool kits discussed here are general-purpose established libraries like IBM's \code{qiskit}\footnote{\url{https://qiskit.org/}} and the niche software stack designed specifically for QNLP, namely \code{lambeq} \cite{kartsaklisLambeqEfficientHighLevel2021}. \code{lambeq} is a Python library (released in October 2021) built for QLNP experiments and is the first of its kind. As such, this option requires a significant time investment in learning about the packages from source code, accepting the risk of running into some bugs, and a relatively limited number of features compared to \code{qiskit}. The alternative option would be to build custom tools from scratch using  \code{qiskit}. This alternative would allow for maximal control over the design but it would require a significant amount of additional work in building the tools already provided by \code{lambeq} and other libraries (e.g., \code{discopy} \cite{de_Felice_2021}). Moreover, \code{lambeq} is naturally integrated with \code{tket} \cite{pytket}, a tool used for cross-platform quantum circuit compilation and optimization so the quantum circuits built with \code{lambeq} can be directly executed on IBM simulators/hardware or converted to \code{qiskit} objects for further modification. Therefore, we decided to use \code{lambeq} so that initial design and experimentation could be done immediately rather than spending significant time on replication.

To build a quantum kernel, we must define an algorithm to compute the fidelity of two quantum states. The two similarity measures considered in this project are the transition amplitude (Fig.~\ref{fig:trans-amp-circ} and Eq.~\eqref{eq:transition-amp-kernel}) and SWAP test (Fig.~\ref{fig:swap-circ} and Eq.~\eqref{eq:swap-test-sim}) approaches. The design that is canonically optimal with respect to noise and accuracy is not immediately clear. 
One might opt for the transition amplitude approach because the states are not arbitrary so the similarity can be estimated directly from the feature maps of each data point, according to Eq.~\eqref{eq:transition-amp-kernel}. Further, this approach requires at least two less qubits to implement (one for the ancillary qubit and one for the shared $s$ qubit(s)), meaning that it minimizes the width of the circuit. Since this approach requires an inversion and composition of the two given circuits it has an added technical challenge of processing the results and determining how to compose the circuits when there are multiple qubits per pregroup type, $q_s>1$. This trade-off between the two fidelity measures in the context of QNLP is investigated further in this work, but since the the transition amplitude approach minimizes width and it computes the overlap in a more direct manner, it will be initially preferred.

\section{Models and Simulation}

\subsection{Initial Training}  \label{sec:task}
% this may not be the proper section heading
% describe how to kernelize discocat - that is, use it as a feature map
% talk about training embeddings

\begin{figure}
    \centering
    \includegraphics[width=\linewidth]{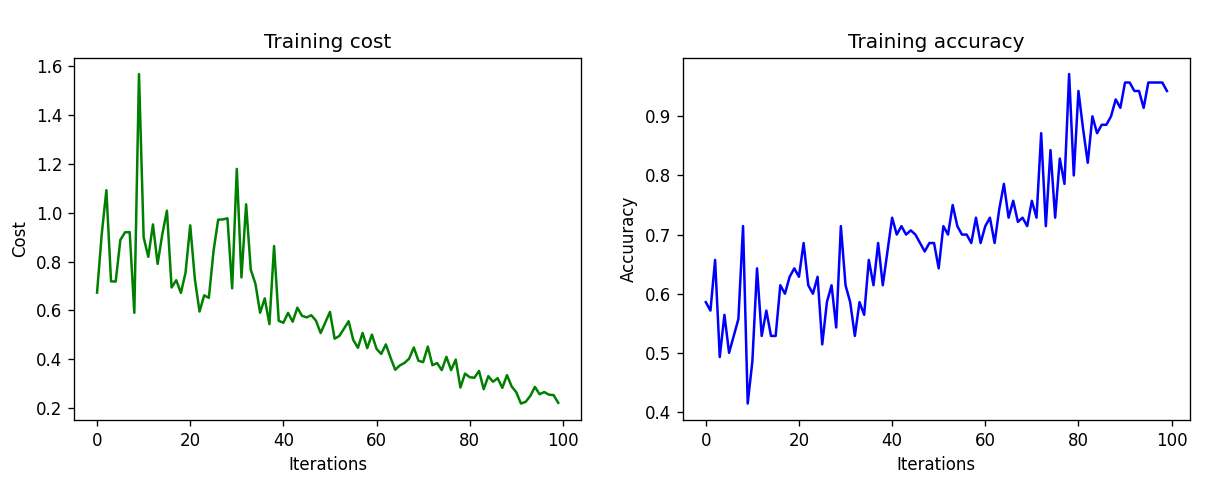}
    \caption[Word embedding training convergence.]{Training convergence for cross-entropy loss (left) and accuracy (right) of the explicit model used to learn word embeddings over 100 epochs. A one layer IQP ansatz with one qubit per sentence and noun types ($q_s=1=q_n$) was used. Circuit was executed on IBM's Aer quantum simulator with no noise for 8192 shots.}
    \label{fig:replication-training}
\end{figure}

We must first determine how to load sentences into their appropriate quantum state. Past work \cite{zengQuantumAlgorithmsCompositional2016, oriordanHybridClassicalquantumWorkflow2020} has relied on qRAM, but since this is infeasible at the moment we adopt the paradigm that uses the DisCoCat model as an envelope to encode our sentences using QML techniques as demonstrated in recent work \cite{lorenzQNLPPracticeRunning2021, meichanetzidisGrammarAwareQuestionAnsweringQuantum2020}. This is analogous to the classical ML technique of learning word embeddings but now has the advantage of incorporating linguistic knowledge as conveyed by DisCoCat. This was done by replicating the meaning classification experiments done in Ref.~\cite{lorenzQNLPPracticeRunning2021}. We used their same dataset and model to learn the quantum word embeddings and benchmark our proposed models. The classification task consists of determining if a given sentence's topic is ``food'' or ``technology'' related. The dataset is built from a vocabulary of size 17 and has 100 sentences which split according to a 70/30 train/test split. Some data points include:
\begin{center}\vspace{-2mm}
\begin{tabular}{c}  \vspace{-2mm}
 ``skillful man prepares tasty meal'', \\  \vspace{-2mm}
 ``skillful person debugs program'', \\  \vspace{-2mm}
 ``woman prepares useful application'',   
\end{tabular}
\end{center}
and other grammatically correct combinations of the vocabulary. While this task is small and relatively simple compared to state-of-the-art NLP work, it is nontrivial since some words are shared between the classes, making it an illustrative task for this proof-of-concept design. Following the work in Ref.~\cite{lorenzQNLPPracticeRunning2021}, we use an \textit{Instantaneous Quantum Polynomial} (IQP) ansatz for the quantum circuit with one layer and one qubit to express the sentence, $s$, and noun, $n$, grammar types -- that is $q_s=1=q_n$. An example of this ansatz is given in Fig.~\ref{fig:discocat-qc}. We then trained the model using appropriate post-selection and the expectation value of the sentence qubit as the label. This explicit model achieved a training accuracy of $95.31 \pm 0.01$\% and testing accuracy of $93.09 \pm 0.01$\%, estimated over seven trials. The training progress is given in Fig.~\ref{fig:replication-training}.
% Train acc: 0.9530612244897959 +/- 0.005562327705055075
% Test acc: 0.930942380952381 +/- 0.007085941039455665

% NOTE: lost the training progress data so I am kind stuck with these plots so I just made them bigger

It  should be noted that due to the inherent probabilistic nature of these models in addition to the limited size of the dataset, the accuracies are subject to fluctuations. We attempted to correct for this by executing circuits for 8192 shots and estimating our accuracies over seven trials using different random number seeds to report both the mean accuracy and its standard error.

\subsection{Transition Amplitude}

Armed with a classification task and appropriate word embedding parameters, we could experiment with the various state fidelity measures available. As mentioned in Section~\ref{sec:meth}, the transition amplitude approach uses less qubits and allows us to directly implement \qcsim using the quantum feature map, $\widehat U_{\phi (\vec x_i)}$, and its adjoint. This suggests that this approach may be more suitable for this task. To implement this, we initially tried the simplistic approach of taking the adjoint of the second data point, composing the two quantum circuits, and executing with post-selecting for 0 on all qubits. However, this performed very poorly, resulting in roughly a 50\% accuracy. Even the similarity of a point with itself was not near a value of one as we would expect. 

The flaw with this design is that it is in effect finding the transition amplitude of the entire circuit, but we only want the transition amplitude of the $s$ qubit(s) (i.e., the bold string in Fig.~\ref{fig:inner-p}). This requires removing post-selection from the $s$ qubit(s) and then finding the probability the this reduced state is 0 after post-selecting the other qubits. This is a more technically challenging post-processing task, especially since it is difficult to keep track of the desired $s$ qubit(s) with the chosen software stack. Thus, this kernel function was only implemented for the case where a single qubit expresses a sentence type, $q_s=1$. Re-normalizing the probabilities after post-selection, we obtain a value between 0 and 1, which directly corresponds to similarity.

\begin{figure}[h]
    \centering
    \includegraphics[width=0.85\linewidth]{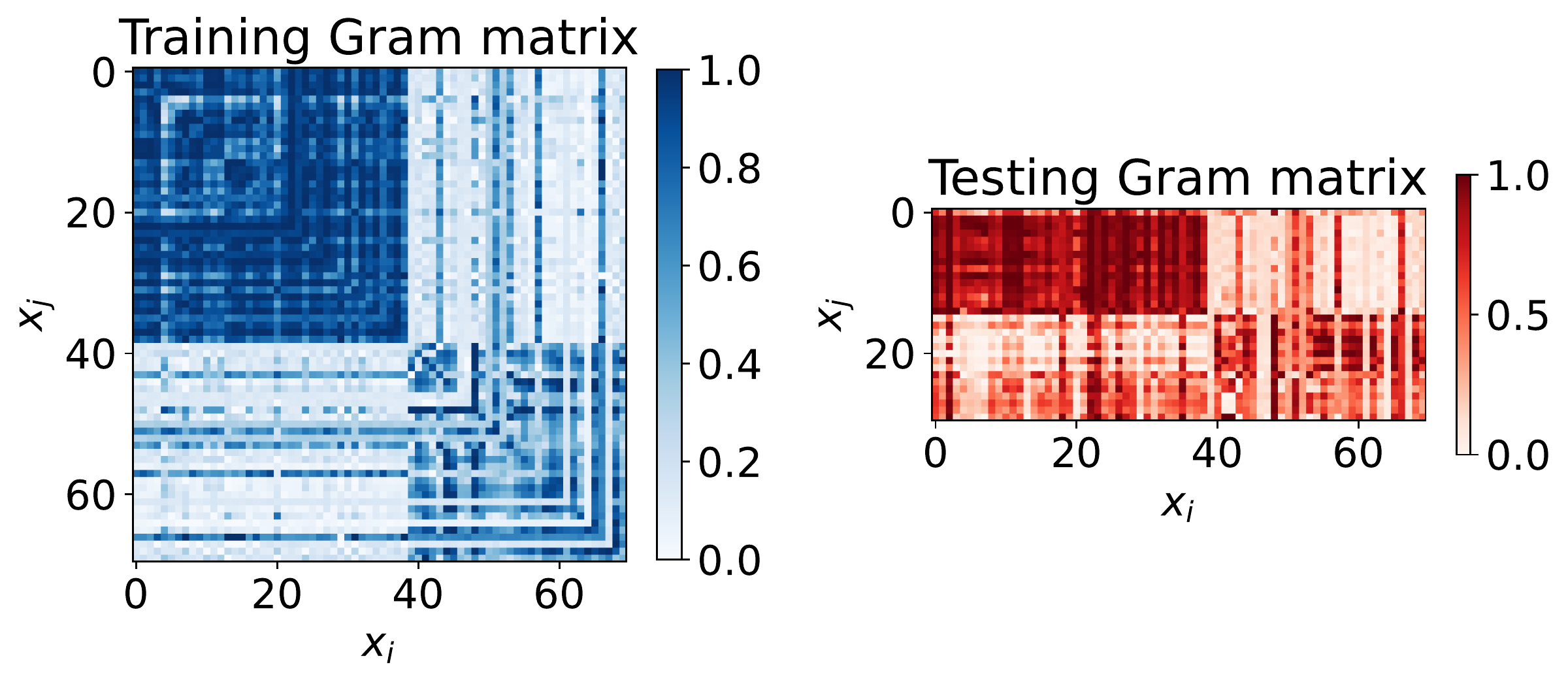}
    \caption[Gram matrices of the transition amplitude approach.]{Gram matrices of the transition amplitude approach for training and testing. These matrices hold elements, $K_{i, j}=|\braket{\phi (\vec x_j) \vert \phi (\vec x_i)}|^2$, which were computed using the approach in Eq.~\eqref{eq:transition-amp-kernel}. An SVM fit to these matrices achieves a training accuracy of 91.4\% and testing accuracy of 96.7\%. Data points are ordered by class label for visualization.}
    \label{fig:trans-amp-gram1}
\end{figure}

\begin{table}[h]
  \centering
  \caption[Summary of similarity matrix regions generated from the transition amplitude approach.]{Summary of similarity values obtained from the transition amplitude approach for different regions of the Gram matrix. This table reports the mean and standard deviation, $\sigma$. The regions studied are the corners in Fig.~\ref{fig:trans-amp-gram1} that correspond to data from Class 0 compared with data from Class 0, similarly for Class 1, and then Class 0 data being compared to Class 1 data (mixed).}
  \label{tab:trans-amp}
  \begin{tabular}{c c c}
    \hline
    Region & Mean Similarity & Similarity $\sigma$  \\
    \hline\hline
    Class 0 & 0.86 & 0.15 \\ 
    Class 1 & 0.49 & 0.29 \\ 
    Mixed & 0.22 & 0.21 \\
    \hline
  \end{tabular}
\end{table}
% Class 1 mean: 0.8556561862866372, 	 std: 0.1469020955231786
% Class 0 mean: 0.4869331366410691, 	 std: 0.2851824344407401
% Class non mean: 0.2210091585563451, 	 std: 0.21320426211707716

% Initial tests were promising. The result for a given data point compared with itself was near 1 but often not exa 
This kernel function was used to create the Gram matrices, $K_{i, j}=|\braket{\phi (\vec x_j) \vert \phi (\vec x_i)}|^2$, for both training and testing data, shown in Fig.~\ref{fig:trans-amp-gram1} and numerically summarized in Tab.~\ref{tab:trans-amp}. Note the symmetry of the kernel function, i.e., $K_{i, j}=K_{j, i}$, allows us to reduce the number of circuit executions for training. A SVM classifier from \code{scipy}\footnote{\url{https://scipy.org/}} was then fit and tested on these precomputed matrices. This resulted in a training accuracy of $89.59 \pm 0.01$\% and testing accuracy of $95.72 \pm 0.01$\%, which is an increase in accuracy compared to the explicit model that was used to train the word embeddings in Section~\ref{sec:task}. 
% Train acc: 0.8959183673469389 +/- 0.009447143875230124
% Test acc: 0.9572380952380952 +/- 0.0056350515777957595

While accuracy was high, the similarity matrices in Fig.~\ref{fig:trans-amp-gram1} are not as we expected. The data points are ordered by class so one would expect to see a four tile chess board plot. Interestingly, the upper left corner (class 0 compared to class 0) of the matrices are more uniformly dark, which indicates high similarity throughout while the bottom right corner (class 1 compared to class 1) is less solid and dark. This is quantified in Tab.~\ref{tab:trans-amp} where we can see that class 0 has a high mean similarity and low variance compared to class 1.
We suspect that the reason for this imbalance is that adjoint of the sentence feature map circuit is not as ``powerful'' of a reversing operation as we expected. That is, class 0 data points were trained to map the $s$ qubit to $\ket{0}$ so these states coming into the adjoint feature map are already close to $\ket{0}$ but the opposite is true for the class 1 data -- the states are close to $\ket{1}$ so the adjoint has to do a more difficult task to rotate the state back to $\ket{0}$. This creates a larger chance of error for class 1 data. Further (and most concerning), the diagonal of the training Gram matrix is not all ones as would be expected for the similarity of a point with itself. This suggests that there is a flaw with the approach but this was not investigated further because of time constraints and the encouraging  accuracy.

% The above reasons are only speculation, so further work is recommended to determine if this circuit inversion is simply in this application due to post-selection or some other reason because these results do not match theoretical predictions. Due to the time constraints of the project, this is out of the scope of the project and we shift our attention to an alternative approach. 

Achieving an acceptable accuracy on a flawed set of Gram matrices, suggests a certain degree of robustness in the kernelized model paradigm. As a result, we decided not to spend additional time trying to generalize the algorithm to handle higher dimensional sentence tensors but rather experiment with an alternative similarity measure that may be more reliable. This single qubit simulation serves as a valid proof-of-concept test for the approach and increasing $q_s$ would only expand the dimension of the space. This is not likely to address the core problem that we suspect lies in the inversion of the circuit but would raise more theoretical consideration regarding the order to compose circuits.

% It is recommend that the embeddings be retrained with the oppopsitde labels to see of this hypothesis is correct

\subsection{SWAP Test}

The SWAP test similarity measure requires more qubits to implement (specifically, $1+q_s$ additional qubits) then the transition amplitude approach but has the advantage of being able to measure the fidelity of two arbitrary quantum states. The implementation of this approach is less complicated and can easily be generalized to higher dimensional sentence tensors by either performing multiple controlled-SWAP gates on the single ancillary qubit or creating additional ancillary qubits, each with their own test. The question still remains of what qubit pairs we target for the similarity with $q_s>1$ though.

Implementation was done by preparing the two given data points' quantum circuits in parallel, applying a SWAP test (see Fig.~\ref{fig:swap-circ}), and post-selecting the appropriate encoding qubits to get the probability distribution of ancillary qubit. Since \code{lambeq} does not have a supported controlled-SWAP gate, we converted to \code{tket} objects once the individual circuits were constructed. Using Eq.~\eqref{eq:swap-test-sim}, we can then estimate \qcsim and build a quantum kernel function.

\begin{figure}[h]
    \centering
    \includegraphics[width=0.85\linewidth]{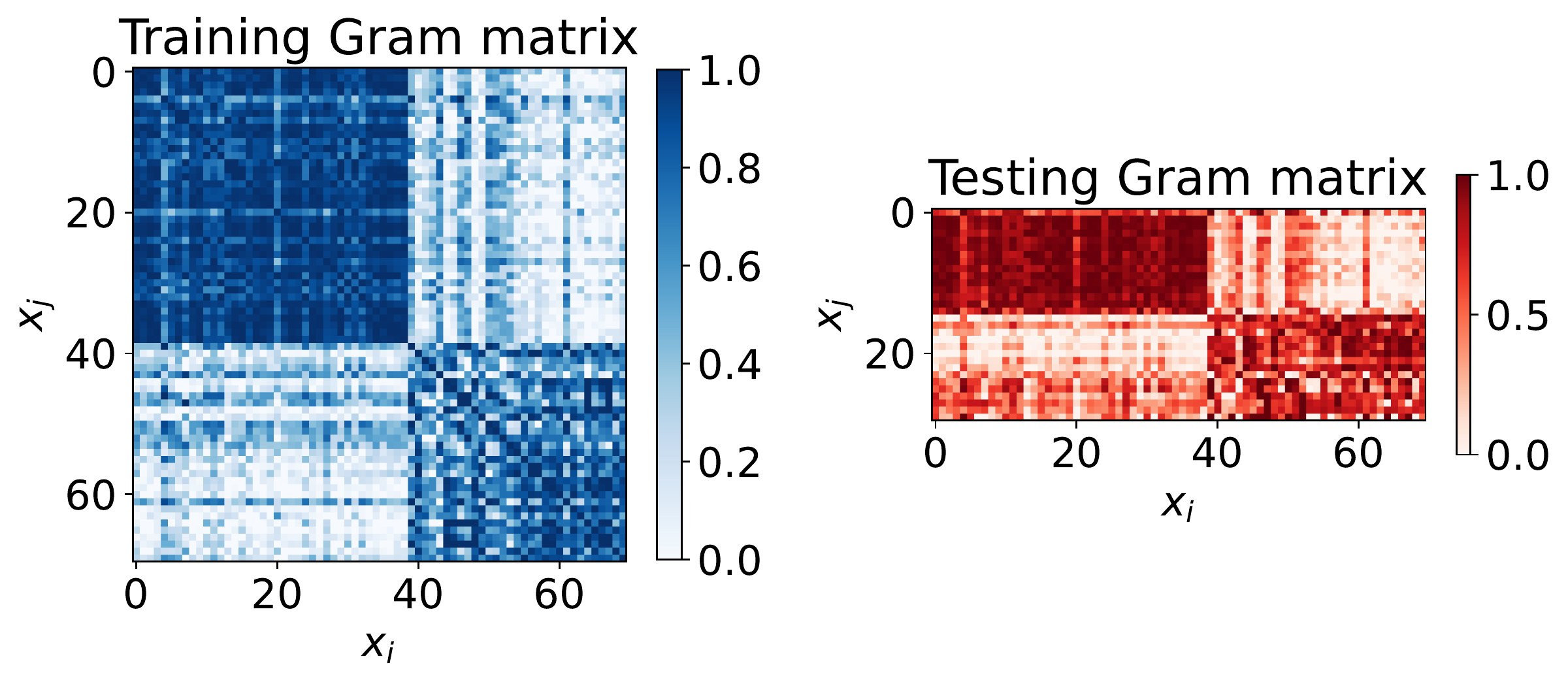}
    \caption[Gram matrices of the SWAP test similarity measure.]{Gram matrices of the SWAP test similarity measure. These matrices hold elements, $K_{i, j}=|\braket{\phi (\vec x_j) \vert \phi (\vec x_i)}|^2$, which were computed using the approach in Eq.~\eqref{eq:swap-test-sim}. An SVM fit to these matrices achieves a training accuracy of 98.6\% and testing accuracy of 100\%. Data points are ordered by class label.} 
    \label{fig:swap-test-gram1}
\end{figure}

\begin{table}[h]
  \centering
  \caption[Summary of similarity matrix regions generated from the SWAP test.]{Summary of similarity values obtained from the SWAP test approach for different regions of the Gram matrix. This table reports the mean and standard deviation, $\sigma$. The regions studied are the corners in Fig.~\ref{fig:swap-test-gram1} that correspond to data from Class 0 comparing with data from Class 0, same for Class 1, and then Class 0 data being compared to Class 1 data (mixed).}
  \label{tab:swap}
  \begin{tabular}{c c c}
    \hline
    Region & Mean Similarity & Similarity $\sigma$  \\
    \hline\hline
    Class 0 & 0.89 & 0.12 \\ 
    Class 1 & 0.68 & 0.27 \\ 
    Mixed & 0.24 & 0.25 \\
    \hline
  \end{tabular}
\end{table}
% Class 1 mean: 0.8933089321655514, 	 std: 0.11652376946883702
% Class 0 mean: 0.678058757757201, 	     std: 0.27206646601999046
% Class non mean: 0.24228411187540624, 	 std: 0.24840985817058003

This kernel function was used to compute the Gram matrices shown in Fig.~\ref{fig:swap-test-gram1} and numerically summarized in Tab.~\ref{tab:swap}. The SVM classifier was then fit and tested on these precomputed matrices. This resulted in a training accuracy of $96.32 \pm 0.01$\% and a testing accuracy of $97.14 \pm 0.01$\%. This outperformed both the explicit model as well as the implicit model built with the transition amplitude approach.  
% Train acc: 0.963265306122449 +/- 0.006984926268636117
% Test acc: 0.9714285714285715 +/- 0.008049088140271584

The Gram matrices in Fig.~\ref{fig:swap-test-gram1} are  closer to what we expect for this task: a more uniform four tile chess board plot with a dark line of all ones along the diagonal. Interestingly, this plot also shows different values of similarity between the two classes, with class 1 again having a slightly lower mean similarity score within itself, relative class 0. This is likely due to the imbalance of the dataset or that the qubits start in $\ket{0}$ but the class 1 mean similarity for the SWAP test is significantly larger than the other approach, scoring 0.68 compared to the transition amplitude's mean score of 0.49.

\subsection{Noise Simulation}
% each approach's gram matrix and acc
% discuss their efficiency
% different training?
% training acc and svm acc

% extensions:
%   noise simulation
%   explore variance

Thus far, we have simulated these quantum similarity measures on an ideal backend from \code{qiskit}. Since current quantum hardware is in the NISQ era and it is not clear when there will be fault-tolerant quantum computers, noise tolerance is a critical factor to consider when dealing with real world applications. Therefore, we simulated our proposed algorithm in the presence of noise to understand the algorithm's robustness.
Recall that the goal of this thesis is the proof-of-concept design of the algorithm so fully dealing with the complexities of decoherence and error was deemed out of the scope of this work.

Using the \code{mock} module of \code{qiskit} we could access simulators of real IBM Quantum hardware. We decided to simulate the \code{ibmq\_guadalupe} device because it has the least number of qubits required for our task and it has the same basis gates our ansatz was designed for \cite{lorenzQNLPPracticeRunning2021}. The explicit model achieved training and testing accuracies of $92.86 \pm 0.01$\% and $91.94 \pm 0.01$\% under the noise simulation. This model performs well in the presence of noise with only a few percentage point drop in accuracy.
% Train acc: 0.9285714285714287 +/- 0.008132499
% Test acc: 0.9194444444444444 +/- 0.0068606977

\begin{figure}[h]
    \centering
    \includegraphics[width=0.85\linewidth]{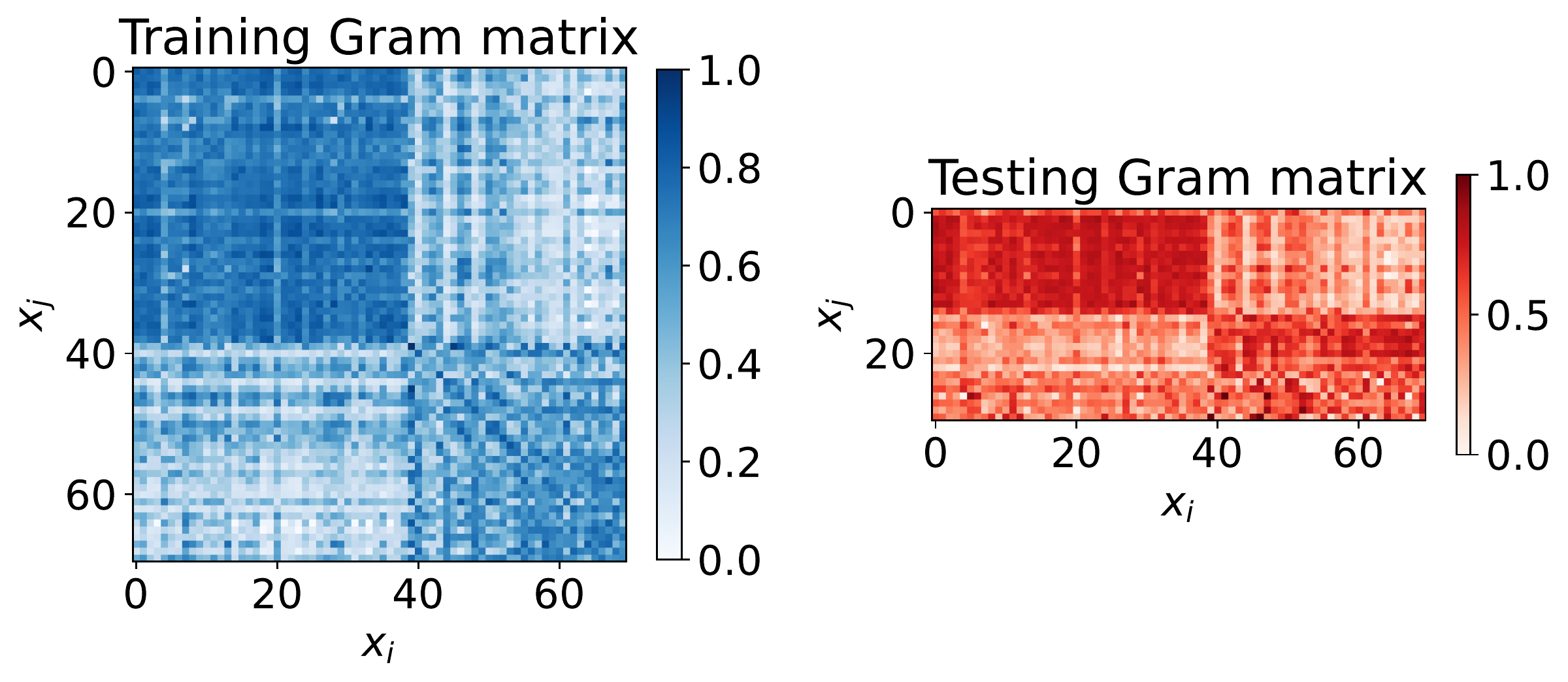}
    \caption[Gram matrices of the SWAP test similarity measure simulated under noise.]{Gram matrices of the SWAP test similarity measure simulated under noise. The noise model was based on the \code{ibmq\_guadalupe} device. Similarity values were computed using Eq.~\eqref{eq:swap-test-sim}. An SVM fit to these matrices achieves a training accuracy of 80\% and testing accuracy of 96\%.} 
    \label{fig:swap-gram-noise}
\end{figure}
% 0.9428571428571428 ; 0.9333333333333333

\begin{table}[h]
  \centering
  \caption[Summary of similarity matrix regions generated from the SWAP test under the presence of noise.]{Summary of similarity values for different regions of the Gram matrix obtained from the SWAP test approach under the presence of noise. This table reports the mean and standard deviation, $\sigma$. The regions studied are the corners in Fig.~\ref{fig:swap-gram-noise} that correspond to data from Class 0 comparing with data from Class 0, same for Class 1, and then Class 0 data being compared to Class 1 data (mixed)}
  \label{tab:swap-noise}
  \begin{tabular}{c c c}
    \hline
    Region & Mean Similarity & Similarity $\sigma$  \\
    \hline\hline
    Class 0 & 0.71 & 0.09 \\ 
    Class 1 & 0.55 & 0.16 \\ 
    Mixed & 0.36 & 0.16 \\
    \hline
  \end{tabular}
\end{table}
% Class 0 mean: 0.7143257113785445, 	 std: 0.09309703342205436
% Class 1 mean: 0.5510294136719404, 	 std: 0.1586650529354587
% Class non mean: 0.36480108364317687, 	 std: 0.16495555928395553

The resulting Gram matrices for the SWAP test simulated under these conditions are shown in Fig.~\ref{fig:swap-gram-noise} and summarized in Tab.~\ref{tab:swap-noise}. This gave accuracies of 80\% and 96.7\% on training and testing data. Note that this simulation was only run once because of its high computational demand so we have no estimate of the error on this accuracy. However, seeding the noiseless simulator with the same value yields respective accuracies of 98.6\% and 100\% and the the similarity matrices in Fig.~\ref{fig:swap-test-gram1} for comparison. The mean similarity scores in each region of the Gram matrices approached a more central value near 0.5 in the presence of noise, which is to be expected. Still, the SVM is able to find the optimal support vectors and classify the testing data with high accuracy.

Unfortunately, the transition amplitude approach could not be tested on the noise simulator. It has not been determined why this is but simulation kills the Python kernel almost immediately so it is likely a memory issue.

\section{Conclusions}
% next steps
%   Investigate circuit inversion
%   Test transition amp approach on 2qubit sentences - how do they compose?
% recommendations

% TODO: NOISE DISCUSSION

The goal of this thesis was to design an implicit quantum kernelized classifier, namely the SVM, that leverages the DisCoCat grammar model. This required designing a function that can measure the similarity of the quantum states of two sentences after the required quantum circuit encoding. We explored two common similarity measures: the transition amplitude approach and the the SWAP test. The implementation and testing of these techniques were done using Python libraries like \code{lambeq} \cite{kartsaklisLambeqEfficientHighLevel2021}, \code{tket} \cite{pytket}, and \code{dicopy} \cite{de_Felice_2021} aided by a \code{qiskit} quantum simulator backend. 
It was shown that the SVM models consistently performed better on the testing dataset than the explicit model that was used to train the embeddings. Between the two similarity measures proposed, the SWAP test approach performed slightly better on both training and testing data, achieving respective accuracies of $96.32 \pm 0.01$\% and $97.14 \pm 0.01$\% compared to the $89.59 \pm 0.01$\% and $95.72 \pm 0.01$\% achieved by the transition amplitude approach. Further, the transition amplitude approach yields a flawed Gram matrix but has the benefit of using less qubits. 
However, we can consider the additional qubits required by the SWAP test  a negligible overhead cost because (i) the number of qubits on real hardware has been steadily growing with no sign of stopping, (ii) the number of qubits required for QNLP tasks is bounded by the effective upper limit on sentence length, and (iii) the additional qubits required by the SWAP test scales as a constant additive factor. Moreover, the SWAP test requires a smaller percentage of post-selection because of the additional qubits but since this will likely introduce additional noise in real experiments this is not a compelling advantage of the SWAP test but is worth noting. The SWAP test was also shown to be robust to noise, achieving a comparable testing accuracy of 96.7\% when simulated under noise, as specified by the \code{ibmq\_guadalupe} device.
Therefore, the increased reliability and performance of the SWAP test and its comparable efficiency, we recommend the SWAP test over the transition amplitude for most applications.

Moving forward, the obvious next step in the analysis would be executing the kernel function on real quantum hardware (ideally a variety of different hardware types) but this requires special access. We also recommend experimenting with more qubits assigned to both noun and sentence grammar types. This will encode the sentence's state over multiple qubits so the correct method for computing similarity between two general sentences is not immediately obvious (i.e., which qubit is compared to which). It will depend on the method used to train the word embeddings, opening up an interesting avenue for further research. Further work could be done to improve the the word embedding scheme. The technique used here is in some sense a variational qRAM that must be trained for each task, vocabulary, and ansatz definition. Finally, an interesting extension to this work would be incorporating the kernel function with the extended model, DisCoCirc \cite{coeckeMathematicsTextStructure2020}, that expresses meaning of an entire text.

\section*{Word Count}
$\approx$~5,600

\FloatBarrier

\pagebreak

\vspace{1cm}
\printbibliography

\pagebreak

\appendix

\section{Statement of work and contributions}\label{sec:AppendixA}
%``The first appendix, Appendix A, must include a statement of any work related to your thesis project that was completed prior to September 1st 2020, and also which clearly describes what work was completed in each of the fall and winter terms.  If some of the work in your thesis was collaboratively done, please state the nature of these collaborations and who did what work."

I was aware of the QNLP field before September 2021 and wanted to design my thesis around the topic so I familiarized myself with the relevant literature in the days leading up to the start of the academic year. Beyond this preliminary reading and ideation, I completed no work relating to this project before the school year. The first semester was spent defining the specific project goal, conducting literature review, and replicating existing results as presented in Section~\ref{sec:task}. The second semester was dedicated to implementation and analysis of the proposed design. 
All code is available on this project's GitHub: \url{https://github.com/mattwright99/thesis}

\end{document}